\documentclass[11pt,a4paper]{article}
\usepackage[hyperref]{eacl2021}

\usepackage{booktabs}
\usepackage{times}
\usepackage{latexsym}
\usepackage{amsmath}
\usepackage{amssymb}
\usepackage{algorithm}
\usepackage[noend]{algpseudocode}
\usepackage{wrapfig}
\usepackage{multirow}
\usepackage[pass]{geometry}
\usepackage{microtype}
\usepackage{todonotes}
\usepackage{lipsum}

\aclfinalcopy % Uncomment this line for the final submission
\def\aclpaperid{862} %  Enter the acl Paper ID here

\algnewcommand{\LineComment}[1]{\State \(\triangleright\) #1}

\title{An Empirical Study on the Generalization Power of Neural Representations Learned via Visual Guessing Games}

\author{Alessandro Suglia$^1$, Yonatan Bisk$^2$, Ioannis Konstas$^1$, Antonio Vergari$^3$, \\\textbf{Emanuele Bastianelli}$^1$, \textbf{Andrea Vanzo}$^1$, \and \textbf{Oliver Lemon$^1$}\\
  $^1$Heriot-Watt University, Edinburgh, UK \\
  $^2$Carnegie Mellon University, Pittsburgh, USA \\
  $^3$University of California, Los Angeles, USA \\
  $^1$\texttt{\{as247,i.konstas,e.bastianelli,a.vanzo,o.lemon\}@hw.ac.uk} \\
  $^2$\texttt{ybisk@cs.cmu.edu},
  $^3$\texttt{aver@cs.ucla.edu} \\}

\date{}

\begin{document}
\maketitle
\begin{abstract}

Guessing games are a prototypical instance of the ``learning by interacting'' paradigm.
This work investigates how well an artificial agent can benefit from playing guessing games when later asked to perform on novel NLP downstream tasks such as Visual Question Answering (VQA).
We propose two ways to exploit playing guessing games: 1) a supervised learning scenario in which the agent learns to mimic successful guessing games and 2) a novel way for an agent to play by itself, called Self-play via Iterated Experience Learning (SPIEL).
We evaluate the ability of both procedures to generalise:
an in-domain evaluation shows an increased accuracy ($+7.79$) compared with competitors on the evaluation suite CompGuessWhat?!;
a transfer evaluation shows improved performance for VQA on the TDIUC dataset in terms of harmonic average accuracy ($+5.31$) thanks to more fine-grained object representations learned via SPIEL. 
\end{abstract}

\section{Background \& Related Work}

Learning a language requires interacting with both the environment and other agents~\cite{bisk2020experience}. Language games represent one common example of this~\cite{wittgenstein1953philosophische}, as seen by the important role of play in L1 child language acquisition~\cite{hainey2016systematic} as well as L2 learners~\cite{godwin2014games}. 

Among the language games defined in the literature~\cite{steels2015talking}, \textit{guessing games} represent the first step in a curriculum for language learning. For example, in GuessWhat?!~\cite{de2017guesswhat}, two agents interact with each other: a \textit{Questioner} generates questions aimed at finding a hidden object in the scene and an \textit{Oracle}, aware of the target object, answers the questions supporting the Questioner in playing the game. Different from other language games~\cite{das2017visual}, guessing games have a specific goal which represents a clear incentive for learning. In addition, they require that the \textit{Questioner} masters both natural language generation and understanding with a focus on object categories and attributes. 
For humans, concepts learned in this way are generic and generalisable to new tasks and domains where grounded reasoning is important~\cite{hampton1979polymorphous}. However, \textit{how well can AI agents generalise with concepts acquired from visual guessing games?}

The literature has not explored if representations built from self-play are transferable, focusing instead on large scale self-supervised learning. 
For instance, large scale image captioning datasets have been used to train multi-modal Transformers
~\cite{lu2019vilbert,li2019visualbert,tan2019lxmert,chen2019uniter}. 
Multi-task learning~\cite{lu202012} has been used to leverage the diversity of training signals provided combining datasets, but only for discriminative tasks. 
While some dialogue work \cite{cogswell2020dialog} aims to bootstrap a conversing agent from VQA datasets, most work on GuessWhat?!~\cite{de2017guesswhat,shekhar2019beyond,strub2017end} has designed bespoke models for the task, ignoring the utility of this dataset for other Vision+Language tasks.

We propose self-play as a mechanism for learning general grounded representations. We seed our approach with the \textit{GuessWhat?!} corpus of questions and objects, and demonstrate how to generalise to other downstream tasks. We propose two different strategies to exploit these data. First, a supervised learning phase is undertaken to learn a Questioner and Oracle model able to play guessing games. Second, the trained agents can be used to play guessing games on images requiring only object annotations as supervision.
We show that an agent trained on GuessWhat?! dialogues can use self-play to adapt to new and harder tasks. Specifically, we investigate models' generalisation performance and quality of the learned representations on the CompGuessWhat?! benchmark~\cite{suglia-etal-2020-compguesswhat}, a more extensive evaluation suite for GuessWhat?!. Furthermore, we study how the learned representation help solve VQA on the dataset TDIUC~\cite{kafle2017tdiuc}. We show overall comparable performance with state-of-the-art models and improvements for specific question types that require object attribute information to be answered correctly. 

\section{Methodology}\label{sec:methodology}
Our proposed transfer/fine-tuning procedure requires a training set of guessing games $D_g$ from which we learn a Questioner $Q$ and an Oracle $O$ via supervised learning. Given a set of images $\mathcal{I}$, it is possible to use the trained models $Q$ and $O$ to run the self-play procedure for $n$ epochs obtaining the model $Q^n$. Finally, given a downstream task $t$ and an associated dataset $\mathcal{D}_t$ based on images from $\mathcal{I}$, we use $Q^n$'s parameters as initialisation for the training procedure on $\mathcal{D}_t$.  

To apply this procedure, both the \emph{Questioner} and the \emph{Oracle} require a \textit{multi-modal encoder} $\Gamma$ able to generate $d$-dimensional representations for the textual tokens $\mathbf{h}_t$, for the objects $\mathbf{h}_o$, as well as fusing the visual and textual modalities in a representation of the current context $\mathbf{h}_c$. After the self-play procedure, only the encoder $\Gamma$ of the model $Q^n$ is used in the fine-tuning process on the downstream task $t$ using the dataset $\mathcal{D}_t$. It is important to underline that the presented self-play procedure does not depend on a specific implementation of the multi-modal encoder $\Gamma$. A possible implementation is presented in Section~\ref{sec:vlp_implementation} and it is used in the experimental evaluation of this paper. 
\begin{figure*}[h]
\centering
\includegraphics[width=0.95\textwidth]{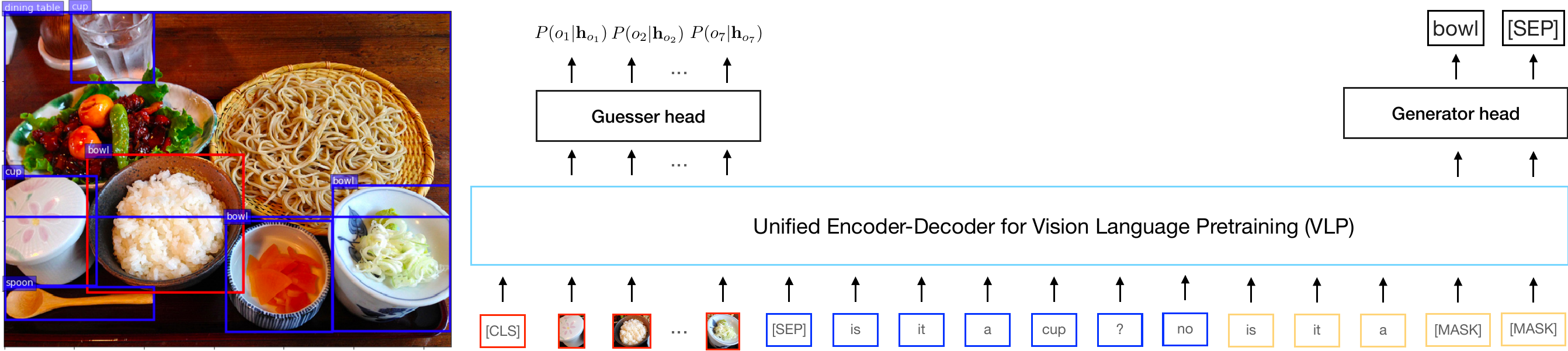}
\caption{We use the single-stream VLP model as a backbone multi-modal encoder for our task. The visual features tokens (marked in red) are the FastRCNN features associated with the objects in the image, the history tokens (marked in blue) and the tokens to be generated (marked in yellow) are given in input to the model. A \textit{Guesser head} uses the learned contextual object representations to generate a probability distribution over the objects $P(o_i | \mathbf{h}_{o_i})$, whereas the \textit{Generator head} is used to incrementally predict the masked tokens.}
\label{fig:multitask_vlp}
\end{figure*}
\subsection{Oracle design}

The Oracle task is cast as a Visual Question Answering (VQA) task conditioned on the image $I$, the current question $q$ and on the target object $\hat{o}$. We follow common practice in vocabulary-based VQA~\cite{antol2015vqa} and we treat the problem as a multi-class classification task over the classes $\{Yes, No, N/A\}$. We use $\mathbf{h}_c$ as input to a multi-layer feedforward neural network to obtain a probability distribution over the label set.

\subsection{Questioner design}

The Questioner must play two roles: question generation and target object prediction~\cite{de2017guesswhat}. It is beneficial to jointly learn the two tasks because the representations learned by each task are complementary. In addition, they better encode attributes, which favours better generalisation to unseen object categories~\cite{suglia-etal-2020-compguesswhat}. 

To solve the two specific tasks in a multi-task fashion, we design two different heads on top of the shared encoder $\Gamma$: 
1) the \textit{guesser head}, produces a probability distribution over every object $o_i$ using the encoded representations $\mathbf{h}_{o_i}$ passed through an MLP; 
2) the \textit{generator head}, a multi-modal decoder, also implemented as an MLP, % multi-layer feed-forward network 
which predicts a probability distribution over the vocabulary $V$ given the context representation generated by $\Gamma$.

We include two losses in our model: 1) the negative log-likelihood of the probability associated by the guesser head with the target object $\hat{o}$~\cite{shekhar2019beyond}; 2) a sequence-to-sequence cross-entropy loss~\cite{sutskever2014sequence} for the generated question tokens.
Unlike previous work that trains a separate module to learn to stop~\cite{shekhar2018ask}, we add a special token \texttt{[STOP]} to the input data so that it learns when to stop more efficiently as part of the question generation task.

Training an agent to solve tasks of different complexity and size is challenging. 
The procedure presented in \cite{shekhar2019beyond} alternates between tasks, updating the hardest task more often. For this technique, finding the right schedule is cumbersome and requires fine-tuning. We rely on a more systematic training procedure based on random dataset-proportional batch sampling inspired by \cite{sanh2019hierarchical}. This represents a hard-parameter sharing multi-task training procedure that avoids interference between tasks and favours a more stable training, which  mitigates catastrophic forgetting~\cite{french1999catastrophic}. 

\subsection{Self-Play via Iterated Experience Learning (SPIEL)}

Inspired by iterated learning ~\cite{kirby2014iterated}, we design a process by which the Questioner learns from games previously generated by other instances of the Questioner agent. We call our training procedure \textit{Self-play via Iterated Experience Learning} (SPIEL).

\begin{algorithm}
\footnotesize
\caption{SPIEL: Self-Play via Iterated Experience Learning}\label{alg:self_play}
\begin{algorithmic}[1]
\Procedure{self\_play}{$Q_0, O, \mathcal{I}, n$}
    \State $\mathcal{D}_q\gets \Call{read\_gold\_games}{\null}$
    \State $\mathcal{E}_g\gets []$ \Comment{Initialise the experience buffer}
    \For{$e\gets 1, n$}
        \LineComment{Interactive phase}
        \State $Q\gets Q^e$ \Comment{load latest weights}
        \State $\mathcal{G}^e\gets \Call{generate\_games}{\mathcal{I}}$
        \State $\mathcal{G}^e\gets \Call{play\_games}{Q, O, \mathcal{G}^e}$
        \State \Call{append}{$\mathcal{E}_g, \mathcal{G}^e$}
        \State $\mathcal{D}^e_g\gets []$ 
                
        \LineComment{Transmission phase}
        \For{$i\gets0, len(\mathcal{E}_g)$}
            \State $g\gets \mathcal{E}_g[i]$ \Comment{Priority to the latest games}
            \If{\Call{is\_valid\_game}{$g$}}
                \State \Call{append}{$\mathcal{D}^e_g$, $g$}
            \EndIf
            \If{\Call{Len}{$\mathcal{D}_g^e$} == \Call{Len}{$\mathcal{D}_q$}}
                \textbf{break}
            \EndIf
        \EndFor
                        
        \LineComment{Learning phase}
        \State $Q^{e+1}\gets \Call{train}{Q, \mathcal{D}_q, \mathcal{D}^e_g}$
    \EndFor
\EndProcedure
\end{algorithmic}
\end{algorithm}

In SPIEL, described in Algorithm~\ref{alg:self_play}, we assume access to a set of images $\mathcal{I}$ and the bounding boxes $\mathcal{O}_I$ of the objects therein.\footnote{Object annotations intended as either gold bounding boxes or predicted bounding boxes from an object detector.} In every gameplay, there is a Questioner $Q$ and an Oracle $O$, initialised with agents $Q^0$ and $O$, respectively, that were trained with Supervised Learning using gold successful dialogues.\footnote{The Oracle is fixed during this learning procedure.} We consider every iteration $e$ of the algorithm as a \textit{self-play} epoch. In a single self-play epoch, we alternate 3 phases: 

\paragraph{Interactive} phase: the agents play guessing games with novel combinations of image and target object. The generated dialogue can be successful if the predicted target object is equal to the target object. Every played dialogue is stored in an experience buffer $\mathcal{E}_g$.

\paragraph{Transmission} phase: in this phase the datasets for the multi-task learning procedure for the Questioner are created. The generator head dataset $\mathcal{D}_q$ is fixed in advance while the dataset for the guesser head $\mathcal{D}_g^e$ is created from the experience buffer $\mathcal{E}_g$ by selecting the \emph{unique} and \emph{valid} dialogues.

\paragraph{Learning} phase: the same multi-task learning procedure used in the supervised learning phase is used to fine-tune the Questioner parameters using the datasets $\mathcal{D}_g^e$ and $\mathcal{D}_q$ collected for the current epoch $e$. This procedure is repeated $n$ times or until a halting condition is reached (e.g. early stopping based on validation metric). 

See Appendix~\ref{appendix:self_play} for implementation details. At the end of the SPIEL procedure, we obtain the model $Q^n$ whose parameters can be reused in other tasks. Particularly, we use the parameters of $Q^n$'s shared encoder $\Gamma$ as initialisation for the fine-tuning on the downstream task $t$ using dataset $\mathcal{D}_t$. 

\subsection{Implementation}\label{sec:vlp_implementation}

We implement a shared multi-modal encoder $\Gamma$ using VLP~\cite{zhou2020vlp}, a single-stream multi-modal Transformer for captioning depicted in Figure~\ref{fig:multitask_vlp}.
During the GuessWhat?! fine-tuning, we extend VLP by including dialogue context in the input together with the features associated with the objects in the image. We learn two new segment ids to represent the question/answer exchanges in the dialogue, as described in \cite{wolf2019transfertransfo}. 
The question is generated by incrementally replacing \texttt{[MASK]} tokens until the end of sequence is generated. 
See Appendix~\ref{appendix:vlp} for more details.
SPIEL training is run on a set of images $\mathcal{I}$ from GuessWhat?! and TDIUC dataset with corresponding object annotations. We make sure that GuessWhat?! test images are not contained in $\mathcal{I}$. This is not an issue for TDIUC test images because the downstream task annotations (QA pairs) are not used by the model during this phase. Once the model has been trained with SPIEL, we use the parameters of the shared encoder $\Gamma$ as a backbone for a VQA model that is fine-tuned on the TDIUC dataset. 

\section{Experimental Evaluation}
To assess the generality of our learned representations, we include two evaluation paradigms:
1) \emph{in-domain evaluation} and 2) \emph{transfer evaluation}. We evaluate several variants of our model: 1) \texttt{VLP-SL}: VLP-based model trained on GuessWhat?! data using multi-task learning; 2) \texttt{SPIEL-gs}: \texttt{VLP-SL} model fine-tuned with our SPIEL procedure where the generator head uses only gold successful games (gs); 3) \texttt{SPIEL-gm}: same as 2) but both successful and failed gold games are used by the generator head. In both SPIEL variants, the guesser head is trained using failed and successful generated games because it is important for the guesser head to be exposed to both types of signal to learn a more robust policy. We decided to investigate the two variants \texttt{SPIEL-gs} and \texttt{SPIEL-gm} to get more insights about the effect that successful and failed games have on the generator head ability to produce effective dialogues.

\begin{figure*}[h]
\centering
\includegraphics[width=0.91\textwidth]{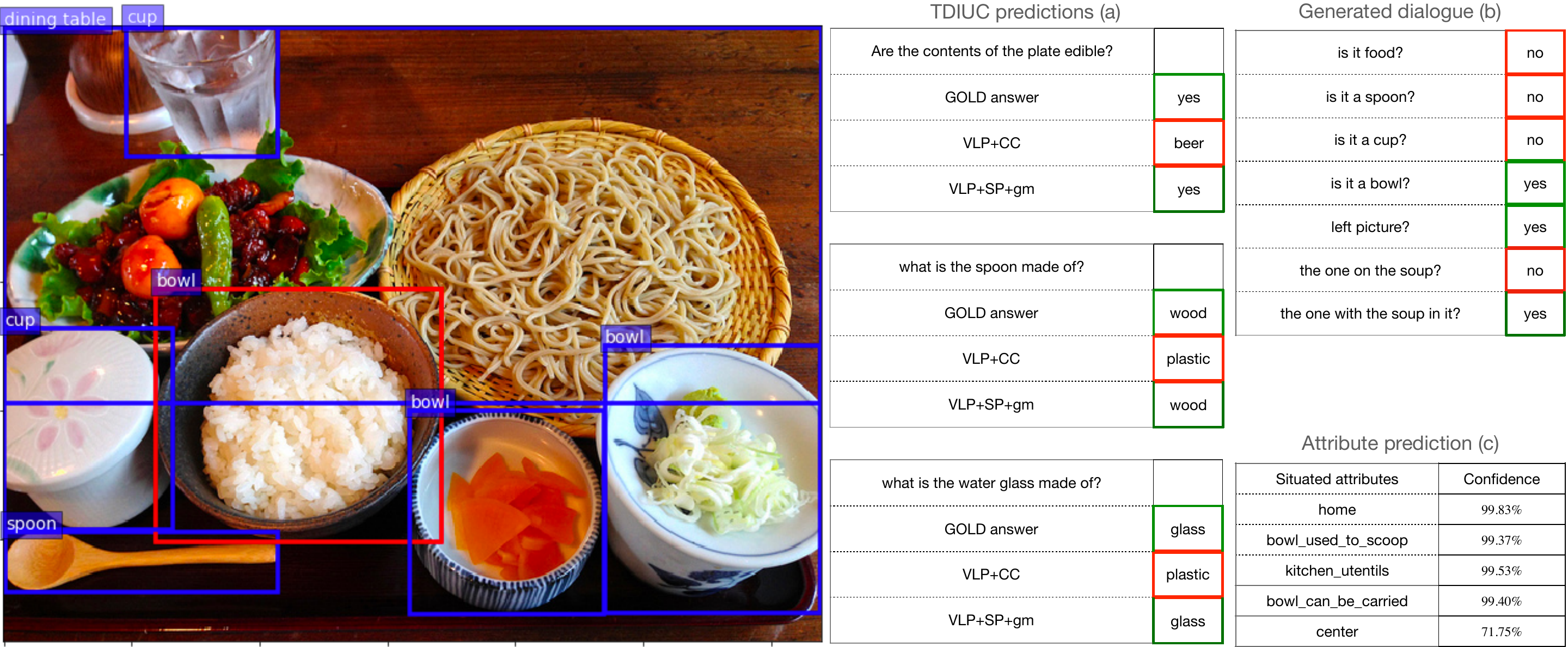}
\caption{We show the ability of the model to play guessing games with the \emph{bowl} as target object (highlighted in \emph{red}). Given the generated dialogue, we use the probing classifier trained for CompGuessWhat?! to predict the \emph{bowl}'s attributes. Predictions on TDIUC questions associated with the current image are reported as well.}
\label{fig:multi_prediction}
\end{figure*}

\subsection{In-domain evaluation}

We use the CompGuessWhat?! evaluation suite~\cite{suglia-etal-2020-compguesswhat} to assess the ability of the Questioner to play guessing games and learn visually grounded representations in the process.
It complements an evaluation based only on gameplay accuracy~\cite{de2017guesswhat} with 2 auxiliary tasks: target object 1) attribute-prediction expressed in terms of abstract attributes (A), situated-attributes (SO), abstract+situated attributes (AS), and location attributes (L); 2) zero-shot gameplay with near-domain accuracy (ND) and out-of-domain accuracy (OD).  Table~\ref{tab:cgw_results} shows the comparison with previous state-of-the-art models on this benchmark such as \citet{de2017guesswhat} (\texttt{DV-*}) and \citet{shekhar2019beyond} (\texttt{GDSE-*}). \texttt{VLP-SL} has a greater advantage in terms of representation power compared to previous models. This is reflected in all the tasks of the CompGuessWhat?! evaluation.
Particularly, we see better performance even for the zero-shot gameplay (ND: $+5.6$, OD: $+15.2$). This is because VLP associates a vector of probabilities that represents a distribution over the VisualGenome object classes with every object. This helps VLP to cope with the issue of unseen objects and helps the model to generalise. Learning to play is key to gameplay performance, leading to an increase of $+4.4$ over \texttt{VLP-SL} and $+7.9$ over \texttt{GDSE-CL}. In this setup, the difference between the versions \texttt{SPIEL-gs} and \texttt{SPIEL-gm} is very minimal ($0.1$). However, when analysed in more detail, we can see that training the questioner with gold successful data only improves attribute prediction while using mixed data improves overall generalisation in the zero-shot evaluation. 

\begin{table}[]
\begin{footnotesize}
\centering
\begin{tabular}{@{}l@{\hspace{4pt}}c
        @{\hspace{8pt}}c@{\hspace{4pt}}c@{\hspace{4pt}}c@{\hspace{4pt}}c
        @{\hspace{5pt}}l@{\hspace{5pt}}
        c@{\hspace{4pt}}c@{\hspace{4pt}}c@{}}
\toprule
     &  & \multicolumn{4}{@{}c@{\hspace{5pt}}}{Attribute Pred.} &  
     &     \multicolumn{2}{@{}c@{\hspace{3pt}}}{ZShot} & Score \\
     \cmidrule{3-6}
     \cmidrule{8-9}
Models              & Acc.  & A     & SO            & AS           & L             & & ND     & OD    &      \\ \midrule
\texttt{Random}              & 15.8  & 15.1  &\phantom{0}0.1 &\phantom{0}7.8&\phantom{0}2.8 & & 16.8   & 18.6  & 13.3 \\
\texttt{DV-SL}          & 41.5  & 46.8  & 39.1          & 48.5         & 42.7          & & 31.3   & 28.4  & 38.5 \\
\texttt{DV-RL}          & 53.5  & 45.2  & 38.9          & 47.2         & 43.5          & & 43.9   & 38.7  & 46.2 \\
\texttt{GDSE-SL}             & 49.1  & 59.9  & 47.6          & 60.1         & 48.3          & & 29.8   & 22.3  & 43.0 \\
%\hspace{1em}+ image & 43.8  & 56.2  & 47.4          & 57.2         & 51.7          & & 39.2   & 39.9  & 45.5 \\
\texttt{GDSE-CL}             & 59.8  & 59.5  & 47.6          & 59.8         & 48.1          & & 43.4   & 29.8  & 50.1 \\
%\hspace{1em}+ image & 52.0  & 57.6  & 47.6          & 58.3         & 50.4          & & 46.6   & 47.0  & 50.7 \\ 
\midrule
\texttt{VLP-SL}              & 59.5  & 59.2  & 48.2          & 59.7         & 49.3          & & 49.0   & 45.0  & 53.5 \\
\texttt{SPIEL-gs}           & 64.1  & \textbf{61.3}  & \textbf{49.6}          & \textbf{61.6}         & \textbf{51.1}          & & 54.9   & 51.9  & 57.8 \\
\texttt{SPIEL-gm}           & \textbf{64.6}  & 60.8  & 48.3          & 59.5         & 51.0          & & \textbf{55.3}   & \textbf{52.9}  & \textbf{57.9} \\ 
\bottomrule
\end{tabular}
\end{footnotesize}
\caption{F1 scores for attribute prediction and accuracies for zero-shot evaluation on CompGuessWhat?!.}
\label{tab:cgw_results}
\end{table}

\subsection{Transfer evaluation}

For the transfer evaluation, we use the VQA dataset TDIUC~\cite{kafle2017tdiuc}.
It provides a finer-grained way to assess the quality of the representations learned by our guessing game transfer technique
in terms of several question types including object categories and their attributes.
Specifically, we were interested in improving on the following question types: 1) Positional reasoning; 2) Counting; 3) Object presence; 4) Utility/Affordances; 5) Attribute; 6) Color; and 7) Object recognition. TDIUC is evaluated using the arithmetic mean accuracy per question type (A-MPT), as well as the harmonic mean (H-MPT) that better captures the skewed question-type distribution.
In Table~\ref{tab:tdiuc_results}, we report a comparison between variants trained on guessing games data (\texttt{VLP+SL} and \texttt{SPIEL-*}), the original model VLP trained on Conceptual Captions (\texttt{VLP+CC}) and other state-of-the-art models \textit{specifically} designed for the VQA task such as \texttt{MUREL}~\cite{cadene2019murel}, \texttt{RAU}~\cite{noh2016rau}, \texttt{NMN}~\cite{andreas2016neural}, \texttt{MCB-*}~\cite{fukui2016multimodal}. The full set of results is available in the Appendix, Table~\ref{tab:full_tdiuc_results}.

Among 
% the reported VQA models, 
them,
\texttt{MUREL} achieves the best scores across the board, due to a custom iterative reasoning mechanism and a non-linear fusion module.
%
% Despite high values of A-MPT, the H-MPT is way lower than all the other models for the VQA-only category. 
%
However, all our models have a more balanced overall performance which results in better harmonic means (H-MPT, $+5$ points over \texttt{MUREL}). 
Specifically, this improvement is favoured by an increase in accuracy on the \emph{Utility/Affordances} question type ($+20.7$). As shown by the attribute prediction in the CompGuessWhat?! and depicted in Figure~\ref{fig:multi_prediction} (c), our models learn better representations than competitors specifically for abstract attributes among which there are object affordances. Particularly, we can see how it is able to understand that certain objects can \emph{contain} things (e.g. ``the one with the soup in it?"), that objects have specific functions (e.g. ``are the contents of the plate edible?") or that they have specific properties (e.g. ``a spoon is made of wood").

The effectiveness of the proposed fine-tuning procedure is confirmed by the improved performance across all the question types compared to our baseline \texttt{VLP+CC}. Models such as \texttt{MUREL} and \texttt{MCB-*} equipped with specific VQA modules have an advantage on specific question (e.g., positional reasoning) compared to VLP that relies only on BERT self-attention layers~\cite{devlin2019bert}. In addition, when comparing the two SPIEL variants, a similar trend showed in the in-domain evaluation can be observed. Particularly, \texttt{SPIEL-gm} benefits from being exposed to more language data coming from successful and failed guessing games. 
 
\newcommand{\SH}[1]{\begin{scriptsize}#1\end{scriptsize}}
\begin{table}[!t]
\centering
\begin{footnotesize}
\begin{tabular}{@{}l@{\hspace{2pt}}c@{\hspace{3pt}}c@{\hspace{3pt}}c@{\hspace{3pt}}c@{\hspace{3pt}}c@{\hspace{3pt}}c@{\hspace{3pt}}c@{\hspace{3pt}}|@{\hspace{3pt}}c@{\hspace{3pt}}c@{\hspace{3pt}}c@{}}
\textsc{Model}  & {\rotatebox{65}{\textsc{\SH{Position}}}} & {\rotatebox{65}{\textsc{\SH{Count}}}} & {\rotatebox{65}{\textsc{\SH{Presence}}}} & {\rotatebox{65}{\textsc{\SH{Afford.}}}} & {\rotatebox{65}{\SH{\textsc{Attr.}}}} & {\rotatebox{65}{\textsc{\SH{Color}}}} & {\rotatebox{65}{\textsc{\SH{Recog.}}}} &  \rotatebox{65}{\textsc{\SH{A-MPT}}}               & \rotatebox{65}{\textsc{\SH{H-MPT}}}               \\ 
\toprule
\texttt{RAU}        & 35.3         & 48.4         & 94.4         & 31.6         & 56.5         & 66.9         & 86.1        & 67.8         & 59.0        \\
\texttt{NMN}        & 27.9         & 49.2         & 92.5         & 25.2         & 47.7         & 54.9         & 82.0        &  62.6         & 51.9        \\
\texttt{MCB-A}      & 55.4         & 51.0         & 93.6         &\textbf{35.1} & 56.7         & 68.5         & 81.9        & 67.9         & \textbf{60.5}        \\
\texttt{MCB}        & 33.3         & 50.3         & 91.8         & 33.9         & 53.2         & 56.9         & 84.6        & 65.8         & 58.0        \\ 
\texttt{MUREL}      & \textbf{41.2}& \textbf{61.8}& \textbf{95.8}& 21.4         & \textbf{58.2}& \textbf{74.4}&\textbf{89.4}& \textbf{71.2}& 59.3        \\
\midrule% \midrule
% \toprule
\texttt{VLP}      & & & & & & & & & &\\
\hspace{5pt}+\texttt{CC}     & 36.9         & 55.3         & 94.7         & 31.0         & 55.4         & 67.3         & 85.8        & 68.8         & 60.1        \\
\hspace{5pt}+\texttt{SL}  & 39.0         &\textbf{57.6} &\textbf{94.8} & \textbf{42.1}& 54.3         & 69.0         & 86.1        & 70.5         & 64.0        \\
\texttt{SPIEL-gs}  &\textbf{40.9} & 57.5         &\textbf{94.8} & 36.3         & 56.9         & 69.2         &\textbf{86.3}& 70.4         & 63.3        \\
\texttt{SPIEL-gm}  & 40.6         & 57.0         &\textbf{94.8} & 39.2         &\textbf{57.0} &\textbf{69.4} & 86.2        & \textbf{70.9}         &\textbf{64.3}\\ 
\bottomrule
\end{tabular}
\end{footnotesize}
\caption{Results for the transfer evaluation on TDIUC. The models are divided in two categories: (top) Models specifically designed for VQA and (bottom) our VLP-based implementations. We report only the question types that we believe will benefit from the guessing games fine-tuning procedure. For the full set of results please refer to Appendix, Table~\ref{tab:full_tdiuc_results}.
}
\label{tab:tdiuc_results}
\end{table}

\section{Conclusions}

In this work, we verified that representations learned while playing guessing games can be transferred to other downstream tasks such as VQA. We presented two ways of learning from guessing games data namely multi-task learning and SPIEL. Models using SPIEL performed better both on in-domain evaluation on CompGuessWhat?! as well as on the transfer task TDIUC. Our self-play procedure was able to learn useful and finer-grained object representations such as object affordances, thus demonstrating that \textit{learning to guess helps learning to ground}.

The current study showed how we can apply the SPIEL training procedure to a VQA dataset such as TDIUC. We believe that this work can be extended to other datasets because the SPIEL procedure only requires a set of images and associated object bounding boxes. These could be either gold or generated by a trained object detector therefore classifying guessing games as a \textit{holistic self-training procedure} for multi-modal datasets.

\bibliography{eacl2021}
\bibliographystyle{acl_natbib}

\appendix
\section{Appendices}
\label{sec:appendix}

\subsection{Self-Play via Iterated Experience Learning (SPIEL)} \label{appendix:self_play}

Learning to replicate gold dialogues is not enough to play successfully. High performance in gameplay can be achieved only when the agents start playing the game and are exposed to their own mistakes. Reinforcement Learning~\cite{strub2017end} or Collaborative Learning~\cite{shekhar2019beyond} are possible approaches to tackle this problem. 

Inspired by iterated learning ~\cite{kirby2014iterated}, we design a process by which ``the gameplay arises in one instance of the questioner through induction on the basis of observations of gameplay in other questioner agents who acquired that gameplay capability in the same way". Therefore, we call our procedure \textit{Self-play via Iterated Experience Learning} (SPIEL).

In this setup, we assume we have access to a set of images $\mathcal{I}$ and for each image $I$ we have object bounding boxes $\mathcal{O}_I$. The SP training procedure, showed in Figure~\ref{alg:self_play}, can be described as follows. We assume that there is a Questioner agent $Q$ and an Oracle agent $O$. At the beginning of the procedure they are initialised with agents $Q^0$ and $O$, respectively, trained with Supervised Learning using gold successful dialogues~\footnote{The Oracle is fixed during this learning procedure.}. We consider every iteration $e$ of the algorithm as a \textit{self-play} epoch. In a single self-play epoch we alternate 3 phases: 1) \textit{interactive} phase: the agents play guessing games with novel combinations of image and target object; 2) \textit{transmission} phase: the questioner creates new datasets from the dialogues generated over the epochs; 3) \textit{learning} phase: multi-task learning is used to fine-tune the Questioner parameters using the datasets collected for the current epoch.

\subsubsection{Interactive phase}

We start the interactive phase by first sampling a set of reference games $\mathcal{G}^e$ which consists of pairs $(I, \hat{o})$ where $I \in \mathcal{I}$ and $\hat{o}$ is the target object sampled at random from the object annotations $\mathcal{O}_I$. The agents $Q^e$ and $O$ play the games $\mathcal{G}^e$ and accumulate the generated experiences. During this phase, the questioner agent is using the most updated weights generated at epoch $e-1$. It generates questions by nucleus sampling~\cite{holtzman2019curious} from the probability distribution over the vocabulary learned by the generator head. When the \texttt{[STOP]} token is sampled, the guesser head, conditioned on the dialogue generated so far, selects the object $\tilde{o}$ with the highest probability. A game is \emph{successful} if the predicted object $\tilde{o}$ is equal to the target object $\hat{o}$. 

\subsubsection{Transmission phase}

For every epoch $e$, in the transmission phase, we create the datasets $\mathcal{D}_q$ and $\mathcal{D}_g$ for the questioner and guesser heads, respectively, used in the learning phase for the questioner parameters update.

\noindent \textbf{Questioner experience buffer}
To make sure that the questioner does not experience \textit{language drift}~\cite{lee2019countering}, we consider a fixed dataset $\mathcal{D}_q$ composed of dialogues generated by humans contained in the GuessWhat?! training data. The shared encoder $\Gamma$ benefits from this data too because it is still exposed to human generated language, which  guarantees better generalisation.

\noindent \textbf{Guesser experience buffer}
The Guesser should learn from its own mistakes -- therefore we use generated dialogues for the model updates~\cite{de2017guesswhat, shekhar2019beyond}. Inspired by Prioritised Experience Replay~\cite{schaul2015prioritized}, we create the experience buffer for the guesser $\mathcal{E}_g^e$ by accumulating all the \emph{unique} and \emph{valid} dialogues generated until epoch $e$. We consider a dialogue \emph{unique} if $\mathcal{D}_g^e$ does not contain another dialogue with the same encoding~\footnote{The encoding of a dialogue is the SHA-256 hash associated with its sequence of tokens.}. In addition, we consider a dialogue \emph{valid} if it does not contain repeated questions. We cap the number of dialogues in $\mathcal{D}_g^e$ so that it matches the number of experiences in $\mathcal{D}_q$. This is done so that during the multi-task training procedure there is an equal number of dialogues for each task from which the agent will learn.

\subsubsection{Learning phase}

In this phase, we use the same multi-task training procedure that was used during the supervised learning phase. We update the Questioner parameters using the dialogues collected in $\mathcal{D}_q$ and $\mathcal{D}_g^e$. The updated parameters resulting from this step will be used for the self-play epoch $e+1$.

\subsection{VLP implementation} \label{appendix:vlp}

\subsubsection{Multi-modal encoder}

To implement the agents in our guessing games, we rely on VLP, a single-stream multi-modal model~\cite{zhou2020vlp} that jointly learns visual and language representations using Conceptual Captions (CC) dataset~\cite{sharma2018conceptual}. The input starts with a classification token (\texttt{[CLS]}), followed by a series of $K$ visual tokens, a separation token (\texttt{[SEP]}) divides the \emph{dialogue} sequence from the visual and from the sequence of \textit{tokens to be generated}.  
In a guessing game, we represent the reference image $I$ as a set of image regions extracted from an off-the-shelf object detector $\{r_1, r_2, \dots, r_K\}$. Following \cite{zhou2020vlp}, each region $r_i$ is represented by linear transformation of a feature vector $f \in \mathbb{R}^{d_n}$, region class probabilities $c \in \mathbb{R}^{d_c}$ and region geometric information $g \in \mathbb{R}^{d_o}$ where $d_o = 5$ consists of four values for top left and bottom right corner coordinates of the region bounding box (normalized between 0 and 1) and one value for its relative area (i.e., ratio of the bounding box area to the image area, also between 0 and 1). The Questioner models uses at most 36 predicted bounding boxes from FastRCNN while the Guesser is using features generated by FastRCNN for gold bounding boxes. We use a specific segment id $s_v$ for every region.

For the language part, we use Wordpiece embeddings~\cite{wu2016google}. In particular, we  flatten the turns of the dialogue context as a sequence of tokens. However, to allow the model to differentiate between question and answer tokens, following \cite{wolf2019transfertransfo}, we rely on novel segment ids ($s_u$,$s_a$). The VLP's hidden state of the \texttt{[CLS]} token is used as context representation $\mathbf{h}_c$.

\subsubsection{Oracle design}

The implementation of the Oracle follows the one presented in the original VLP paper to solve the VQA task~\cite{zhou2020vlp}. Particularly, the model predicts a probability distribution over the possible answers by using a multi-layer feed-forward neural network that receives in input the element-wise product between the hidden state associated with the \texttt{[CLS]} token and the hidden state associated with target object. The model is optimised by minimising the cross-entropy loss using as training dataset the question/answer pairs in the successful GuessWhat?! training dialogues. 

\subsubsection{Questioner design}

We rely on the VLP ability to generate captions for the question generation task. In particular, we provide in input to the model: 1) predicted FastRCNN visual features following~\cite{zhou2020vlp}; 2) dialogue generated so far as a flattened sequence of tokens; 3) question to be generated. We use another segment id $s_q$ to allow the model to differentiate what is the input and which are the tokens to be generated. Following \cite{dong2019unilm}, we make sure that the attention mask for tokens of the question to be generated are masked so that the token at timestep $t$ is not allowed to attend to the future tokens (\emph{seq2seq attention mask}). For this specific model, we use the masked language modelling objective~\cite{devlin2019bert} casting the task as multi-modal masked language modelling.
\begin{table}[t]
\centering
\begin{small}
\begin{tabular}{@{}ll@{}}
\toprule
Model              & Accuracy \\ \midrule
Human              & 90.80\%  \\
Random             & 17.10\%  \\
LSTM               & 61.30\%  \\
HRED               & 61\%     \\
LSTM+VGG           & 60.50\%  \\
HRED+VGG           & 60.40\%  \\
\midrule
ParallelAttention & 63.40\%  \\
GDSE-SL            & 62.96\%  \\
GDSE-CL            & 59.79\%  \\ \midrule
VILBERT            & 65.69\%  \\
VLP-SL             & 69.30\%  \\
SPIEL-gs          & \textbf{71.80\%}  \\
SPIEL-gm          & 71.70\%  \\ \bottomrule
\end{tabular}
\end{small}
\caption{Results for the guesser accuracy evaluation on gold dialogues.}
\label{tab:gw_guesser}
\end{table}

\subsection{GuessWhat?! evaluation} \label{appendix:guesswhat}

\paragraph{Oracle evaluation} We report the test accuracy for the Oracle of 82.22\%. The baseline model used by all the other is $78.5\%$~\cite{de2017guesswhat}.

\paragraph{Guesser evaluation} We report in Table~\ref{tab:gw_guesser} the accuracy of the guesser in predicting the target object when gold dialogues are given in input. We compare this model with several baselines reported in \cite{de2017guesswhat} (first block), more sophisticated methods such as \texttt{ParallelAttention}~\cite{zhuang2018parallel} and \texttt{GDSE-*}~\cite{shekhar2019beyond} (second block) as well as other Transformer-based models such as VILBERT~\cite{lu202012} (third block).

\begin{table*}
\footnotesize
\centering
\resizebox{\textwidth}{!}{% <------ Don't forget this %
\begin{tabular}{@{}llllllllllllllll@{}}
\toprule
\textsc{Model}  & {\rotatebox{90}{\textsc{\underline{Positional}}}} & {\rotatebox{90}{\textsc{\underline{Counting}}}} & {\rotatebox{90}{\textsc{\underline{Presence}}}} & {\rotatebox{90}{\textsc{\underline{Affordances}}}} & {\rotatebox{90}{\textsc{\underline{Attribute}}}} & {\rotatebox{90}{\textsc{\underline{Color}}}} & {\rotatebox{90}{\textsc{\underline{Recognition}}}} & 
\rotatebox{90}{\textsc{Scene}} & 
\rotatebox{90}{\textsc{Absurd}} & \rotatebox{90}{\textsc{Sentiment}} & \rotatebox{90}{\textsc{Activity}} & \rotatebox{90}{\textsc{Sport}} & \rotatebox{90}{\textsc{Accuracy}}                & \rotatebox{90}{\textsc{A-MPT}}               & \rotatebox{90}{\textsc{H-MPT}}               \\ \midrule
MUREL           & 41.19                     & 61.78                   & 95.75                   & 21.43                              & 58.19                    & 74.43                & 89.41                      & 96.11          & 99.80           & 60.65              & 63.83             & 96.20          & {\color[HTML]{333333} 88.20} & {\color[HTML]{333333} 71.20} & 59.30                        \\
RAU             & 35.26                     & 48.43                   & 94.38                   & 31.58                              & 56.49                    & 66.86                & 86.11                      & 93.96          & 96.08           & 60.09              & 51.60             & 93.47          & 84.26                        & 67.81                        & 59.00                        \\
NMN             & 27.92                     & 49.21                   & 92.50                   & 25.15                              & 47.66                    & 54.91                & 82.02                      & 91.88          & 87.51           & 58.02              & 44.26             & 89.99          & 79.56                        & 62.59                        & 51.87                        \\
MCB-A           & 55.40                     & 51.01                   & 93.64                   & 35.09                              & 56.72                    & 68.54                & 85.54                      & 93.06          & 84.82           & 66.25              & 52.35             & 92.77          & 81.86                        & 67.90                        & \textbf{60.47} \\
MCB             & 33.34                     & 50.29                   & 91.84                   & 33.92                              & 53.24                    & 56.93                & 84.63                      & 92.04          & 83.44           & 65.46              & 51.42             & 92.47          & 79.20                        & 65.75                        & 58.03                        \\ \midrule
VLP-CC          & 36.93                     & 55.28                   & 94.65                   & 30.99                              & 55.42                    & 67.33                & 85.76                      & 92.98          & 98.34           & 62.62              & 51.34             & 94.11          & 85.60                        & 68.81                        & 60.14                        \\
VLP-SL   & 39.04                     & 57.61                   & 94.79                   & 42.11                              & 54.29                    & 69.01                & 86.07                      & 93.39          & 97.54           & 65.77              & 52.39             & 94.34          & 85.98                        & 70.53                        & 63.95                        \\
SPIEL-gs & 40.94                     & 57.53                   & 94.76                   & 36.26                              & 56.87                    & 69.2                 & 86.33                      & 93.97          & 97.48           & 62.3               & 54.44             & 94.62          & \textbf{86.1}  & 70.39                        & 63.34                        \\
SPIEL-gm & 40.6                      & 57.01                   & 94.77                   & 39.18                              & 56.97                    & 69.42                & 86.21                      & 93.72          & 97.19           & 66.09              & 55.29             & 94.18          & 86                           & \textbf{70.89} & \textbf{64.31} \\ \bottomrule
\end{tabular} %
}
\caption{Summary of results for the transfer evaluation on TDIUC. The models are divided in two categories: (1) Models which are specifically designed for VQA (top) and (2) models that rely on the VLP encoder to generalise to different downstream tasks (bottom). We underline the question types that we believe will benefit from the guessing games transfer/fine-tuning procedure.}
\label{tab:full_tdiuc_results}
\end{table*}

\end{document}